\def\year{2020}\relax
\documentclass[letterpaper]{article} 
\usepackage{aaai20}  
\usepackage{times}  
\usepackage{helvet} 
\usepackage{courier}  
\usepackage[hyphens]{url}  
\usepackage{graphicx} 
\usepackage{amsmath}
\usepackage{amssymb}
\usepackage{mathrsfs}
\usepackage{multirow}

\usepackage{graphicx}  
\nocopyright
\title{Linear Context Transform Block}
\author{Dongsheng Ruan\textsuperscript{\rm 1,2,3}, Jun Wen\textsuperscript{\rm 1,2}, Nenggan Zheng\textsuperscript{\rm 1}\thanks{Corresponding author}, Min Zheng\textsuperscript{\rm 3}\\
\textsuperscript{\rm 1}Qiushi Academy for Advanced Studies, Zhejiang University, Hangzhou, Zhejiang, China\\ 
\textsuperscript{\rm 2}College of Computer Science and Technology, Zhejiang University, Hangzhou, Zhejiang, China\\
\textsuperscript{\rm 3}State Key Laboratory for Diagnosis and Treatment of Infectious Diseases, School of Medicine, \\Zhejiang University, Hangzhou,Zhejiang, China\\
\{21530003, junwen, zng, minzheng\}@zju.edu.cn
}
 
\begin{document}

\maketitle

\begin{abstract}
Squeeze-and-Excitation (SE) block presents a channel attention mechanism for modeling global context via explicitly capturing dependencies across channels. However, we are still far from understanding how the SE block works. In this work, we first revisit the SE block, and then present a detailed empirical study of the relationship between global context and attention distribution, based on which we propose a simple yet effective module, called Linear Context Transform (LCT) block. We divide all channels into different groups and normalize the globally aggregated context features within each channel group, reducing the disturbance from irrelevant channels. Through linear transform of the normalized context features, we model global context for each channel independently. The LCT block is extremely lightweight and easy to be plugged into different backbone models while with negligible parameters and computational burden increase. Extensive experiments show that the LCT block outperforms the SE block in image classification task on the ImageNet and object detection/segmentation on the COCO dataset with different backbone models. Moreover, LCT yields consistent performance gains over existing state-of-the-art detection architectures, e.g., 1.5$\sim$1.7\% AP$^{bbox}$ and 1.0\%$\sim$1.2\% AP$^{mask}$ improvements on the COCO benchmark, irrespective of different baseline models of varied capacities. We hope our simple yet effective approach will shed some light on future research of attention-based models.

\end{abstract}
\section{Introduction}
\noindent Attention mechanism has achieved remarkable success in a variety of computer visual tasks, e.g., image classification \cite{wang2017residual,hu2018squeeze-and-excitation}, object detection \cite{wang2018nonlocal,zhang2018progressive}, and semantic segmentation \cite{zhang2018context,li2018pyramid}. The attention module is typically plugged into existing deep networks to improve their representational power \cite{he2016deep,xie2017aggregated,szegedy2015going,zagoruyko2016wide,zhang2018shufflenet,howard2019searching}. One of the most prominent works is the Squeeze-and-Excitation network (SENet) \cite{hu2018squeeze-and-excitation}, which is channel-attention based and aims to selectively emphasize informative channels and suppress trivial ones through explicitly modeling dependencies across channels. SENet achieves significant performance gains across varied models and has been successfully applied to a variety of computer vision tasks \cite{sandler2018mobilenetv2,ma2018shufflenet,yu2018learning,howard2019searching}. 

 \begin{figure}[tb]
 \centering
 \includegraphics[width=80mm]{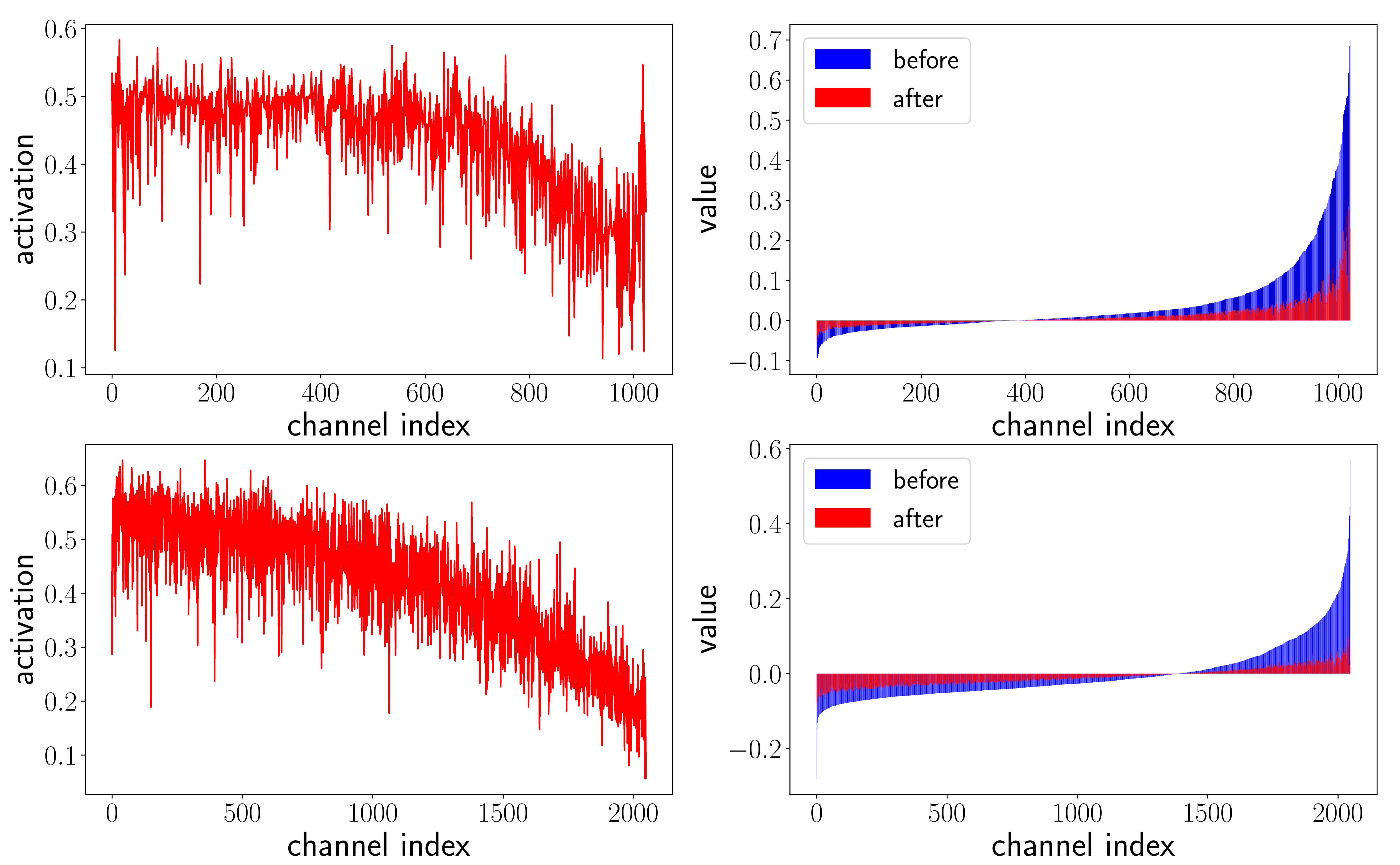}\\
 \caption{Visualization of the averaged attention values (first column) and global context features before and after the first SE block (second column) in stage $4$ (first row) and stage $5$ (second row) on the ImageNet validation set.}\label{fig:attention}
 \end{figure}

 \begin{figure*}[tb]
 \centering
 \includegraphics[width=130mm]{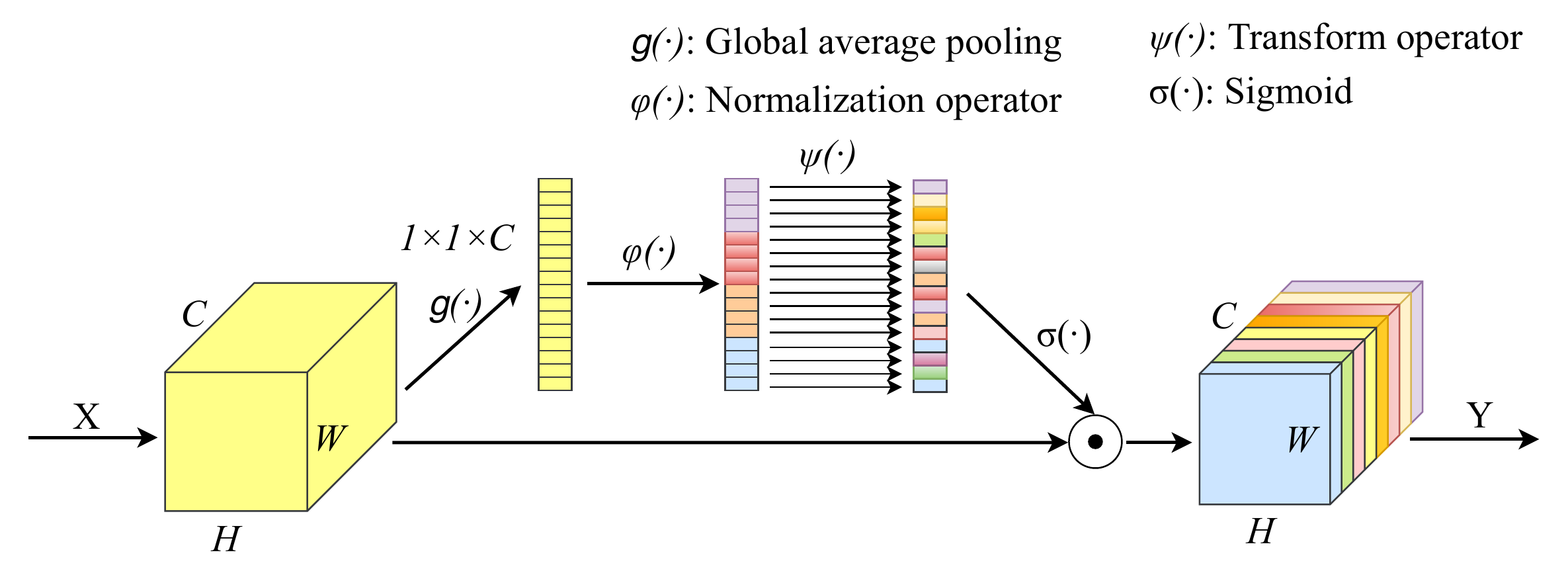}\\
 \caption{Architecture of the linear context transform block. The input feature maps are defined as $\mathbf{X}\in \mathbb{R}^{C\times H\times W}$, where $C$ is the number of channels and $H, W$ are the spatial dimensions. $\mathbf{Y}\in \mathbb{R}^{C\times H\times W}$ denotes the output of the LCT block. $\odot$ denotes broadcast element-wise multiplication.}\label{fig:lct}
 \end{figure*}
 
To dive into this attention mechanism, we are curious of the following two questions: 1) what is the relationship between global context and attention distribution, and 2) which kind of channels are trivial. To answer these questions, we visualize the averaged global context features before and after SE and the corresponding attention activations on the ImageNet validation set. For easier observation, the averaged global context features before SE are sorted in ascending order, as shown in Fig. \ref{fig:attention}. Interestingly, we observe such a negative correlation in SE that global contexts with larger absolute values tend to be attached with smaller attentions, indicating that these channels are generally trivial. By learning such a correlation, SE effectively suppresses these channels and reduces contextual variations across channels, which enables subsequent filters to extract more primitive semantic features and improve generalization ability. Given this observation, a question naturally arises: can we lean such a correlation in a better way?

The SENet has shown the effectiveness of explicit dependency modelling across channels. However, a potential problem for SENet is that when the number of feature channel becomes higher, it will be much more difficult to capture the dependency across all channels to learn such a correlation stably because lots of irrelevant information from other channels can be introduced. A alternative approach is to boost the capacity of context feature transform module as in the GENet \cite{Gather-Excite}, but which brings significant increase of model complexity.

In this paper, we propose a simpler and robuster approach to learning the above negative correlation with a novel module, called Linear Context Transform (LCT) block, which is extremely lightweight and brings negligible parameters and computational burden increase. Specifically, LCT achieves context feature transform with the following two cheap operators: \emph{normalization} and  \emph{transform}. To enable stable context modelling, we divide channels into different groups and normalize the global context within each group using the \emph{normalization} operation. With the \emph{transform} operation, we then linearly transform the normalized global contexts for each channel independently. With varied architectures, we investigate the difference between the SE block and our LCT block in terms of attention distribution and global context feature, and find that the combined  \emph{normalization} and \emph{transform} operators play a similar role as the fully connected (FC) layers of SE in learning the negative correlation while with smaller fluctuations (Fig. \ref{fig:attention_lct_all}). In summary, our main contributions can be summarized as follows:
 \begin{itemize}
    \item We present an empirical study of the relationship between global context and attention distribution of the SENet, and find a negative correlation between them two, which help researchers better understand the mechanism of channel-wise attention and shed light on future research of attention-based models.
     \item We propose a novel light-weight attention block (LCT) for global context modeling by combining simple group normalization and linear transform. To our best knowledge, this is the first work to model global context for each channel independently.
    \item Comprehensive experiments with three visual tasks (image classification on the ImageNet and object detection/segmentation on the COCO) consistently demonstrate the superiority and generalization abiltiy of our attention model.
    \end{itemize}
    
\section{Related Work}
\subsubsection{Normalization}
Batch normalization (BN) \cite{ioffe2015batch} is a milestone technique that normalizes the statistics for each training mini-batch to stabilize the distributions of layer inputs, which enables deep networks to train faster and more stably. However, the property that depends on the mini-batch size leads to a rapid decline in network performance when the batch size becomes smaller.  A series of normalization methods \cite{ba2016layer,ulyanov2016instance,wu2018group,salimans2016weight} have been proposed to address this issue caused by inaccurate batch statistics estimation. Layer normalization (LN) \cite{ba2016layer} computes the statistics along the channel dimension and is well suited for recurrent neural network. Instance normalization \cite{ulyanov2016instance} proposes to perform the normalization across spatial locations. Group normalization (GN) divides features into different groups and normalize them within each group \cite{wu2018group,wen2019exploiting}. Since GN does not exploit the batch dimension, it is still able to achieve high accuracy even in small batch size. 

The design of LCT is inspired by GN. Instead of stabilizing the distribution of layer inputs, LCT is essentially a channel-wise attention mechanism that aims to model global context dependency with group normalization.

\subsubsection{Attention modules}
Recently, several attention modules \cite{chen2017sca,wang2018nonlocal,A2Net,fu2019dual,huang2018ccnet} have been proposed to exploit global contextual information to enhance the representational power of the networks. In particular, SENet \cite{hu2018squeeze-and-excitation} develops a lightweight attention block to recalibrate feature channels by exciting the aggregated contexts from original features. Further, GENet \cite{Gather-Excite} proposes a gather-excite framework for better context exploitation and yields further performance gains at the expense of increasing parameters. GCNet \cite{cao2019gcnet} combines simplified non-local block \cite{wang2018nonlocal} and SE block \cite{hu2018squeeze-and-excitation} to effectively model the global context via addition fusion. In addition to channel attention, CBAM \cite{woo2018cbam} and BAM \cite{park2018bam} exploit both spatial and channel-wise information to yield further performance gains. SKNet \cite{li2019selective} proposes a dynamic selection mechanism that enables the network to adaptively adjust receptive field. More recently, Li \emph{et al.} \cite{li2019spatial} introduce a spatial group-wise enhance module to spatially enhance the semantic expression in each group, showing excellent performance in image classification and object detection.
 
Our work builds on the idea developed in the SE block. However, different from SE, LCT implicitly captures channel-wise dependencies and linearly models the global context of each channel, which is more lightweight and robust.

\section{Method}
In this section, we first review the SE block, and then present the proposed linear context transform (LCT) block.

\subsection{Revisiting the SE block}

The SE aims to emphasize informative features and suppress trivial ones by modeling the channel-wise relationship. To obtain contextual information, SE proposes to squeeze global spatial information. Specifically, it aggregates global context information across spatial dimension through global average pooling operation. Further, to fully capture channel-wise dependencies, the SE block excites the aggregated contexts using two fully-connected layers. Here we define $\mathbf{X} \in \mathbb{R}^{C\times H\times W}$ as the input feature maps of SE, where $C$ is the number of channels and $H, W$ are the spatial dimensions. The SE block can be formulated as:

\begin{equation}
\mathbf{Y} = \mathbf{X} \cdot \sigma (f(g(\mathbf{X})))=\mathbf{X} \cdot \sigma (\mathbf{W}_2ReLU(\mathbf{W}_1g(\mathbf{X}))),
\end{equation}
where $\cdot$ denotes channel-wise multiplication and $g(\mathbf{\cdot})$  global average pooling to generate channel-wise statistics. $\mathbf{W}_1$ and $\mathbf{W}_2$ denote the weights of FC layers and $\sigma(\cdot)$ the $sigmoid$ function.

As shown in Fig. \ref{fig:attention}, the SE performs a non-linear transform to learn a negative correlation between global contexts and attention values by explicitly capturing the dependencies across channels. However, this negative correlation is learned from all channels, which may bring in each channel irrelevant information from other channels and make the global context modeling unstable, resulting in incorrect mapping. To tackle this problem, we propose the novel LCT block.

\subsection{Linear context transform block}
In this section, we introduce the proposed LCT block in detail, which is illustrated in Fig. \ref{fig:lct}. As summarized in GCNet \cite{cao2019gcnet}, global context modeling framework can be abstracted as the following three modules: (a) context aggregation; (b) context feature transform; (c) feature fusion, which framework is also followed by the LCT.

\subsubsection{Context aggregation}
Context aggregation aims to help the network capture long-range dependencies by exploiting information beyond the local receptive fields of each filter. A number of aggregation strategies can be chosen to aggregate contextual information, such as second-order attention pooling \cite{A2Net}, global attention pooling \cite{Gather-Excite,cao2019gcnet}, and global average pooling \cite{hu2018squeeze-and-excitation}. Complex aggregation operators can be used to improve performance of the LCT block, but which are not the focus of our work. Hence we simply employ global average pooling to aggregate the global context features of each sample across spatial dimensions generating a channel descriptor as $\mathbf{z}=\{z_k=\frac{1}{H\times W}\sum_{i=1}^{W}\sum_{j=1}^{H}\mathbf{X}_k(i,j):k \in \{1,...,C\}\}$. 

\subsubsection{Context feature transform}
To effectively and efficiently model the context feature, the LCT introduces a pair of lightweight operators: a normalization operator, which normalizes the global context features within each group, and a transform operator, which takes in the normalized global contexts to produce the importance scores. Specifically, we first divide the descriptor $\mathbf{z}$ into groups and then normalize it within each group along channel dimension. More formally, we define $\mathbf{v}^{i}=\{z_{mi+1},...,z_{m(i+1)}\}$ as the $i$-th local context group, where $i\in \{0,...,G-1\}$ and $G$ are the index and the number of groups, respectively. $m=C/G$ is the number of channels per group. The normalization operator $\varphi$ can be formulated as:
\begin{equation}
\hat{\mathbf{v}}^{i}=\varphi(\mathbf{v}^{i})=\frac{1}{\sigma^{i}}(\mathbf{v}^{i}-\mu ^{i}),
\end{equation}
where $\mu ^{i}$ and $\sigma ^{i}$ are the mean and standard deviation of the $i$-th group, respectively, and can be computed as:
\begin{equation}
\mu ^{i}=\frac{1}{m}\sum_{n \in \mathcal{S}_{i}}z_n,\sigma ^{i}=\sqrt{\frac{1}{m}\sum_{n \in \mathcal{S}_{i}}(z_n-\mu ^{i})^2+ \epsilon.}
\end{equation}
Here $\epsilon$ is a small constant. $\mathcal{S}_{i}$ is the set of the $i$-th group of channel index. 

The normalization operator plays two crucial roles in context feature transform. First, it enables each channel to adjust its own context feature by perceiving context information within each group, implicitly capturing dependencies across channels. Second, it can effectively eliminate the inconsistency of the context feature distribution caused by different samples, which stabilizes the distribution of global context features. 

Next, we define the transform operator to be a function $\psi $: $\mathbb{R}^{C} \rightarrow \mathbb{R}^{C}$ that maps the gathered context features $\hat{\mathbf{z}}$ to the importance scores $\mathbf{a}$, formulated as:
\begin{equation}
\mathbf{a}=\psi (\hat{\mathbf{z}})=\mathbf{w} \cdot \hat{\mathbf{z}}+ \mathbf{b},
\end{equation}
where $\hat{\mathbf{z}}=[\hat{\mathbf{v}}^{0},\hat{\mathbf{v}}^{1},...,\hat{\mathbf{v}}^{G-1}]$. $\mathbf{w}$ and $\mathbf{b}$ are trainable gain and bias parameters of the same dimension as $\hat{\mathbf{z}}$. Note that the transform operator $\psi$ is a channel-wise linear transform, which means that information from other channels is not taken into account in the context transform process. In addition, it only introduces the parameters of $\mathbf{w}$ and $\mathbf{b}$, which are almost negligible compared to the entire network. 
 \begin{figure*}[tb]
 \centering
 \includegraphics[width=150mm]{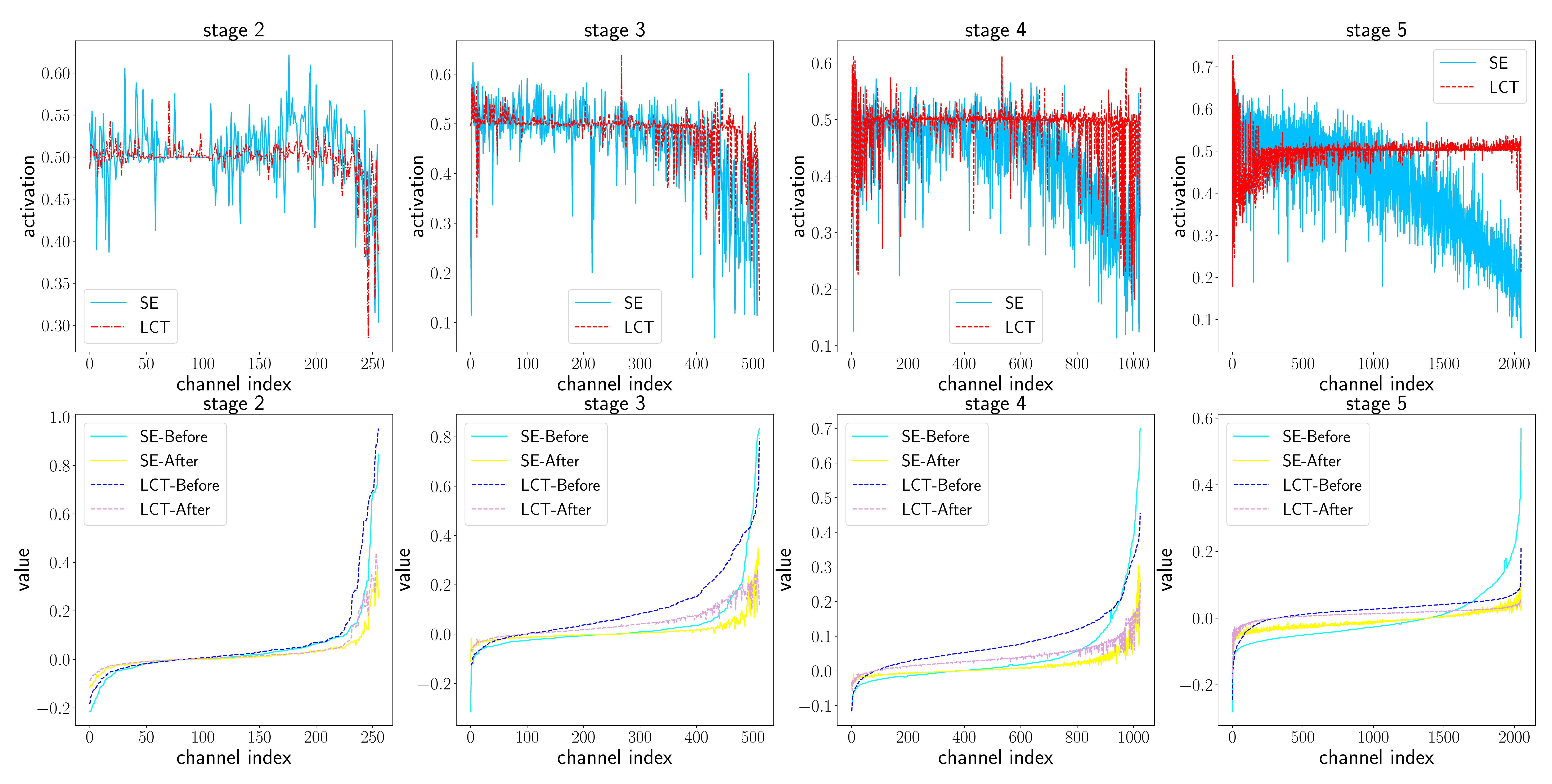}\\
 \caption{Visualizations of the averaged attention values and averaged global context features before and after the first attention blocks at different stages on the ImageNet validation set. The backbone network is ResNet50. Top row: averaged attention valued.  Bottom row: averaged global context features.}\label{fig:attention_lct_all}
 \end{figure*}
 
 \begin{figure}[tb]
 \centering
 \includegraphics[width=70mm]{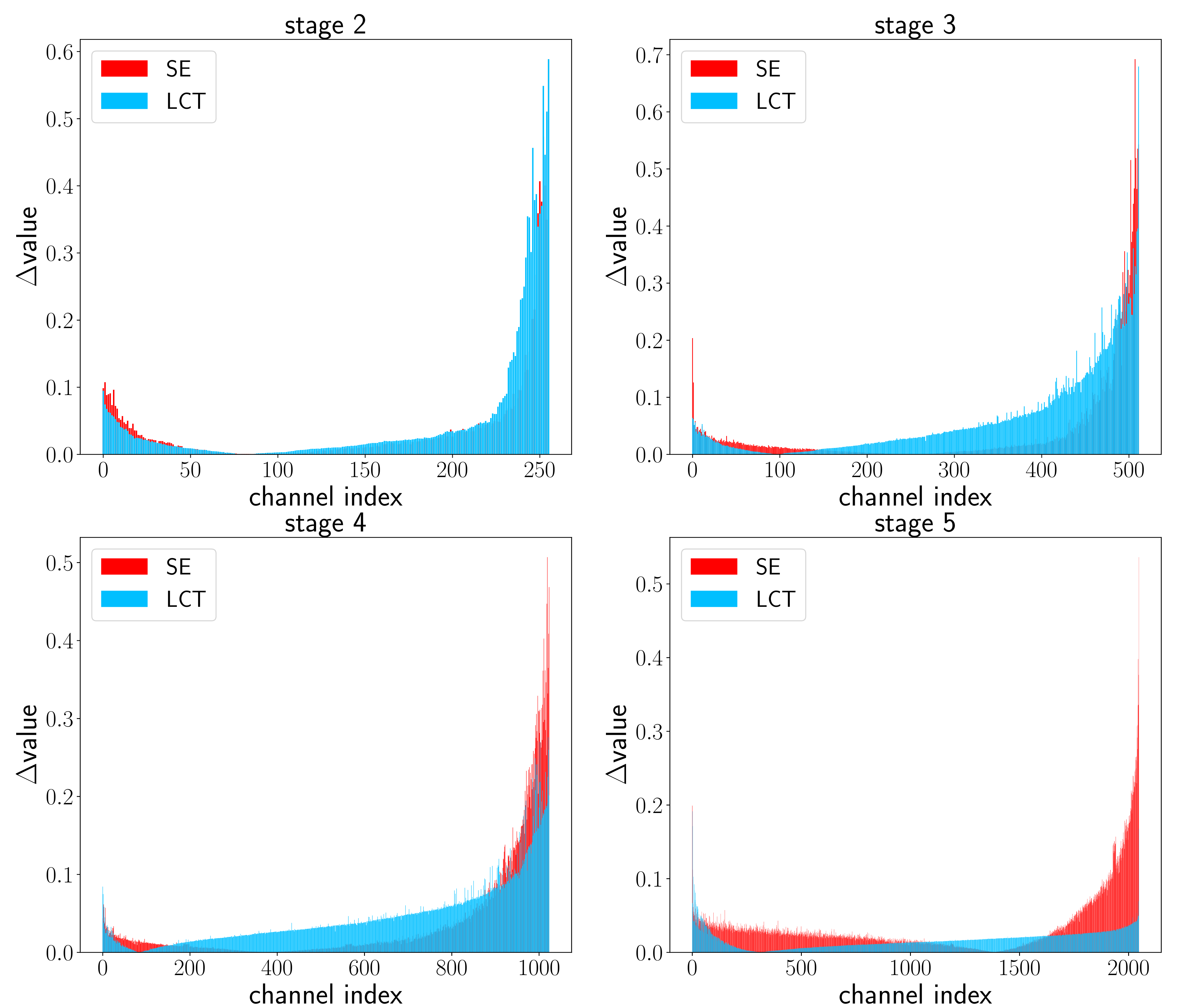}\\
 \caption{Visualizations of the absolute context variations before and after attention blocks at different stages.}\label{fig:right_left}
 \end{figure} 
 
Interestingly, the composition of two operators can be regarded as a special case of GN where the spatial height $H$ and width $W$ are $1$. In the case of $G=1$, it is equivalent to LN. But it is worth noting that the transform operator in the LCT block is designed to transform the global context features, not to compensate for the potential lost of representational ability caused by normalization, which is essentially different from other normalization methods.

\subsubsection{Feature fusion} Finally, the feature fusion module modulates the input features by conditioning on the transformed contexts. Specifically, the output $\mathbf{Y} \in \mathbb{R}^{C\times H\times W}$ of the LCT block is obtained by rescaling the original response $\mathbf{X}$ according to the attention activations $\sigma(\mathbf{a})$ and can be expressed as: 
\begin{equation}
\mathbf{Y}= \mathbf{X} \cdot \sigma(\mathbf{a}).
\end{equation}

\subsubsection{Relationship to SE block}
LCT shares the same context aggregation module and feature fusion module with SE. The main difference between them is the context transform module, which reflects different perspectives of two blocks for global context modeling. First, SE makes use of global information from other channels to help model the global contexts, which actually increases the complexity of context transform. In comparison, our LCT block is more lightweight and simplifies global context modeling by independently transforming the global contexts of each channel. The number of parameters in the SE block is $2C^2/r$, while the number of parameters in the LCT block is $C$, where $r$ is the reduction radio. It is apparent that LCT has significantly decreased parameters. Second, SE explicitly captures channel-wise dependencies using two FC layers, while our approach implicitly captures dependencies within each group through group normalization operator. The results in Table. \ref{tab:group} show that the normalization operator can effectively capture channel dependencies within each group.

\section{Experiments}
In this section, we first evaluate the proposed LCT block on the task of image classification on ImageNet-$1$K \cite{russakovsky2015imagenet}, and then conduct extensive ablation studies. Finally, we experiment on the COCO $2017$ dataset \cite{lin2014microsoft} to demonstrate the general applicability of the LCT block.

\subsection{Image classification on ImageNet}
The ImageNet $2012$ dataset contains $1.28$ million training images and $50$K validation images with $1000$ classes. 

\subsubsection{Implementation details}
We train all models from scratch on $4$ GPUs for $100$ epochs, using synchronous SGD optimizer with a weight decay of $0.0001$ and momentum $0.9$. The initial learning rate is set to $0.1$, and decreases by a factor of $0.1$ every $30$ epochs. The weight initialization is adopted in \cite{he2015delving}. For ResNet50 backbone, the total batch size is set as $256$. For ResNet101 backbone, we reduce the batch size to $220$ due to the limited GPU memory. The standard data augmentation is performed for training: a $224\times224$ crop is randomly sampled from a $256\times256$ image or its horizontal flip using the scale and aspect ratio augmentation. Input images are normalized using the channel means and standard deviations. 

 \begin{table}[tb]
\centering 
\resizebox{1\linewidth}{!}{
\begin{tabular}{l|cccc}  
\hline
Backbone & Params& FLOPs& Top-1 (\%) & Top-5 (\%)\\
\hline
ResNet50&$25.56$M& $4.122$G & $76.15$ & $92.87$\\
+SE & $28.09$M & $4.130$G & $77.31$ & $93.68$ \\ 
+LCT &$25.59$M& $4.127$G &$\mathbf{77.45}$& $\mathbf{93.71}$\\
\hline
ResNet101 & $44.55$M &$7.849$G &$77.37$&$93.56$\\
 +SE & $49.33$M&$7.863$G &$78.49$& $94.19$\\
 +LCT &$44.61$M& $7.858$G &$\mathbf{78.55}$& $\mathbf{94.26}$\\
\hline
\end{tabular}}
\caption{Classification accuracies on the ImageNet validation set. Params denotes the number of parameters. FLOPs denotes the number of multiply-adds.  }
\label{tab:exp-ImageNet}
\end{table}

\begin{table}[t]
\centering 
\resizebox{1\linewidth}{!}{
\begin{tabular}{c|cccccccc}  
\hline
G & $1$& $4$ & $8$& $16$ & $32$ & $64$ & 128\\
\hline
Top-1 &$77.37$&77.36&77.44&$77.34$&$77.32$&$\mathbf{77.45}$&-\\
Top-5 &$93.66$&93.57&93.56&$93.54$&$93.52$&$\mathbf{93.71}$&-\\
\hline
\end{tabular}}
\caption{Classification accuracies (\%) of LCT-ResNet50 with different group numbers $G$ on the ImageNet validation set. - denotes that the network can not converge.}
\label{tab:group}
\end{table}
\begin{table}[t]
\centering 
\begin{tabular}{c|cc}  
\hline
Normalization &  w/ & w/o\\
\hline
Top-1 (\%) &$77.45$&$76.89$\\
Top-5 (\%) &$93.71$&$93.33$\\
\hline
Transform &  w/ & w/o\\
\hline
Top-1 (\%) &$77.45$&$76.82$\\
Top-5 (\%) &$93.71$&$93.32$\\
\hline
\end{tabular}
\caption{Classification accuracies of LCT-ResNet50 with and without normalization/transform operator on the ImageNet validation set.}
\label{tab:norm-trans}
\end{table}

\begin{table}[t]
\centering 
\begin{tabular}{c|ccc}  
\hline
 &  LCT & SE & SE+\\
\hline
Top-1 (\%) &$\mathbf{77.45}$&$77.31$&$77.37$\\
Top-5 (\%) &$93.71$&$93.68$&$\mathbf{93.73}$\\
\hline
\end{tabular}
\caption{Effects of inserting a normalization operator before the two FC layers of the SE block. The backbone is ResNet50. }
\label{tab:SE+}
\end{table}

\begin{table}[t]
\centering 
\begin{tabular}{cc|cc}  
\hline
\multicolumn{4}{c}{Initialization}\\
\hline
$\mathbf{w}$ & $\mathbf{b}$ &Top-1 (\%) & Top-5 (\%)\\
\hline
$0$&$0$&$77.36$&$93.60$\\
$0$& $1$&$\mathbf{77.45}$&$\mathbf{93.71}$\\
$1$&$0$&$77.24$&$93.54$\\
\hline
\end{tabular}
\caption{Results of different initializations with LCT-ResNet50 on the ImageNet validation set.}
\label{tab:init}
\end{table}
As is widely practiced in \cite{hu2018squeeze-and-excitation,woo2018cbam}, our LCT blocks are inserted into each residual block of ResNet. We use $0$ and $1$ to initialize all $\mathbf{w}$ and $\mathbf{b}$ parameters respectively. $G$ is set as $64$ by default. To make a fair comparison, the baseline models are reproduced in the same training settings. We report the top-$1$ and top-$5$ classification accuracy on the single $224 \times 224$ center crop in the validation set.

\subsubsection{Classification results} 
Table \ref{tab:exp-ImageNet} presents the main results of our experiments. We observe that LCT performs better than SE with fewer parameters and less computational burden, irrespective of the depth of the backbone. Compared to ResNet, our LCT block adds few parameters and computations, but achieves significant performance gains ($>1.0\%\uparrow$ on Top-1 accuracy) even in deeper ResNet101. Remarkably, LCT-ResNet50 is able to outperform ResNet101, which indicates that the improvements brought by LCT exceed the benefits of increased network depth ($51$ layers). These results demonstrate effective of LCT on image classification.

\subsubsection{Analysis and discussion} 
To gain some insights into the channel attention mechanism, we investigate the relationship between global context features and attention distribution. Specifically, we first compute the averaged global context features before and after attention blocks and the corresponding attention activations across 1000 classes on ImageNet validation set. Then we sort the averaged global context features in ascending order for better observation. Fig. \ref{fig:attention_lct_all} shows the results of the first attention blocks at different stages. In order to observe the difference more intuitively, we also visualize the $\Delta$value that represents the absolute context variation before and after attention block, shown in Fig. \ref{fig:right_left}.  

We observe that both SE and LCT learn a negative correlation that global context features with larger absolute values tend to be assigned smaller  activations, which suggests that channels with these context features are generally less useful. This is reasonable to some extent, since a large amount of noise is more likely to exist in these channels. When the magnitude of the features of some channels is dramatically larger than that of other channels, subsequent filters will pay more attention on these trivial channels, leading to unstable semantic representation learning. By performing feature recalibration, both blocks effectively suppress the influence of these channels and reduce the contextual differences across channels, which enables subsequent filters to capture robuster semantics of each channel. In a sense, global contexts act like an indicator of which channels need to be suppressed. 

While SE and LCT learn similar attention distributions, there are still several differences. First, the attention distribution learned by LCT is more stable because no other channel information is introduced in the transform operator. Second, LCT  does not over-suppress the original feature responses, thus retaining important semantic information. These findings provide explanations for the effectiveness of the LCT block.


\begin{table*}
\centering 
\resizebox{\textwidth}{!}{
\begin{tabular}{l|l|c|c|ccc|ccc}  
\hline
Detector & Backbone &$\Delta$Params& $\Delta$FLOPs&  AP$_{0.5:0.95}^{bbox}$ & AP$_{0.5}^{bbox} $& AP$_{0.75}^{bbox}$&AP$_{small}^{bbox}$ & AP$_{media}^{bbox} $& AP$_{large}^{bbox}$\\
\hline
\hline
\multirow{3}*{Faster R-CNN} & baseline &-&-&$38.5$ & $60.5$ & $41.8$ & $22.3$ & $43.2$ & $49.8$\\
&+SE& $+4.78$M& $+0.191$G&$39.8$\scriptsize{(+1.3)} & $61.9$ & $43.1$ & $23.9$ & $43.8$ & $\mathbf{51.5}$\\
 &+LCT &$+0.06$M& $+0.187$G & $\mathbf{40.0}$\scriptsize{$\mathbf{(+1.5)}$}& $\mathbf{62.8}$ & $\mathbf{43.4}$ & $\mathbf{24.8}$ & $\mathbf{44.4}$ & $50.9$\\
\hline
\hline
\multirow{3}*{Mask R-CNN}& baseline&-&-&$39.4$ & $61.0$ & $43.3$ & $23.1$ & $43.7$ & $51.3$\\
&+SE &$+4.78$M& $+0.191$G& $40.7$\scriptsize{$(+1.3)$}  & $62.7$ & $44.3$ & $24.5$ & $44.8$ & $52.7$\\
 &+LCT&$+0.06$M& $+0.187$G& $\mathbf{40.9}$\scriptsize{$\mathbf{(+1.5)}$} & $\mathbf{63.1}$ & $\mathbf{44.6}$ & $\mathbf{25.0}$ & $\mathbf{45.1}$ & $\mathbf{52.9}$\\
\hline
\hline
\multirow{3}*{Cascade R-CNN} & baseline&-&-&$42.0$ & $60.3$ & $45.9$ & $23.2$ & $46.0$ & $56.3$\\
&+SE&$+4.78$M& $+0.191$G& $43.4$\scriptsize{$(+1.4)$} & $62.2$ & $47.4$ & $24.7$ & $47.4$ & $57.0$\\
 &+LCT &$+0.06$M& $+0.187$G & $\mathbf{43.6}$\scriptsize{$\mathbf{(+1.6)}$} & $\mathbf{62.4}$ & $\mathbf{47.6}$ & $\mathbf{25.4}$ & $\mathbf{47.6}$ & $57.3$\\
\hline
\hline
\multirow{3}*{Cascade Mask R-CNN } & baseline&-&-&$42.6$ & $60.7$ & $46.7$  & $23.8$ & $46.4$ & $56.9$\\
&+SE& $+4.78$M& $+0.191$G& $43.7$\scriptsize{$(+1.1)$} & $61.8$ & $47.5$ & $24.3$ & $47.5$ & $58.6$\\
 &+LCT&$+0.06$M& $+0.187$G & $\mathbf{44.1}$\scriptsize{$\mathbf{(+1.5)}$} & $\mathbf{62.4}$ & $\mathbf{48.3}$ & $\mathbf{25.0}$ & $\mathbf{47.7}$ & $\mathbf{59.3}$\\
\hline
\end{tabular}}
\caption{Comparisons based on ResNet101 backbone on the task of \textbf{object detection}. $\Delta$Params denotes the change amount of parameters. $\Delta$FLOPs denotes the change amount of computations. The numbers in brackets denote the improvements over the baseline backbone.}
\label{tab:exp-COCO-detec}
\end{table*}

\begin{table*}[tb]
\centering 
\begin{tabular}{l|l|ccc|ccc}  
\hline
Detector & Backbone & AP$_{0.5:0.95}^{mask}$ & AP$_{0.5}^{mask}$& AP$_{0.75}^{mask}$& AP$_{small}^{mask}$ & AP$_{media}^{mask}$& AP$_{large}^{mask}$\\
\hline
\hline
\multirow{3}*{Mask R-CNN } & baseline&$35.9$ & $57.7$ & $38.4$ & $19.2$ & $39.7$ & $49.7$ \\
&+SE & $36.9$\scriptsize{$(+1.0)$}  & $59.4$ & $39.2$ & $20.0$  & $\mathbf{40.8}$ & $50.3$ \\ 
&+LCT &$\mathbf{37.0}$\scriptsize{$\mathbf{(+1.1)}$} & $\mathbf{59.6}$ & $\mathbf{39.3}$&$\mathbf{20.5}$& $\mathbf{40.8}$ & $\mathbf{50.5}$\\
\hline
\hline
\multirow{3}*{Cascade Mask R-CNN} & baseline &$37.0$ & $58.0$ & $39.9$ & $19.1$ & $40.5$ & $51.4$\\
&+SE & $37.7$\scriptsize{$(+0.7)$} & $59.0$ & $40.5$ & $19.4$ & $41.1$ & $52.4$ \\ 
&+LCT&$\mathbf{38.1}$\scriptsize{$\mathbf{(+1.1)}$} & $\mathbf{59.5}$ & $\mathbf{41.3}$& $\mathbf{19.9}$& $\mathbf{41.3}$ & $\mathbf{53.2}$ \\
\hline
\end{tabular}
\caption{Comparisons based on ResNet101 backbone on the task of \textbf{instance segmentation}. The results show that LCT outperforms SE.}
\label{tab:exp-seg}
\end{table*}
\subsection{Ablation study}
\subsubsection{Number of groups}
In this experiment, we assess the effect of group number on the performance of the LCT block. As shown in Table. \ref{tab:group}, LCT is not sensitive to the variation of group number, which is reasonable because the mean and variance do not change significantly with the number of channels per group. We observe that when $G=128$, the network has failed to converge since too many groups may lead to incorrect statistical estimation. When $G = 64$, the performance is slightly higher than that of other settings, indicating that introducing too much information from other irrelevant channels may not be helpful. By default, we set $G=64$ for LCT. Moreover, LCT consistently outperforms SE for all $G$ values, which indicates that the normalization operator can well capture the dependency across channels, even in the extreme case of $G=1$.

\subsubsection{Normalization operator}
To investigate the influence of normalization in the LCT block, we conduct experiments by removing the normalization operator from LCT. Table. \ref{tab:norm-trans} shows the results. It is clear that the LCT block without normalization operator suffers considerable performance degradation. This comparison shows that global context can not be effectively transformed using transform operator alone. It also demonstrates that normalization operator can effectively eliminate the inconsistency of context feature distribution and captures dependencies between channels well. 

We have seen that normalization operator can improve the performance of the LCT block and would like to explore whether normalization operator can also help SE block yield further performance gains. For this purpose, we insert a normalization operator before the FC layers of the SE block. We refer to this block as SE+. $G$ is set to $64$. The results are shown in Table. \ref{tab:SE+}. We find that normalization operator does not bring significant gain to the SE block. The top-$1$ accuracy of the SE+ block is slightly inferior to ours. Based on these results, we can draw the following conclusions: 1) The two FC layers in SE not only can transform the global context features, but also effectively prevent the inconsistency of feature distribution caused by different samples, which is surprisingly similar to two operators in LCT. The difference is that LCT  decomposes the roles of two FC layers into two independent operators, each of which performs its own function. 2) After normalization, a per-channel linear transform is sufficient to transform the global contexts. Introducing information from other channels complicates context feature transform. These findings provide an explanation for the effectiveness of the LCT block.

\subsubsection{Transform operator}

We study the effect of transform operator. To this end, we retain the normalization operator and remove the transform operator from LCT. The results are shown in Table. \ref{tab:norm-trans}. We observe that performance is noticeably reduced and is slightly worse than that without normalization operator, suggesting that transform operator is vitally important for global context transform. The reason is that normalization operator can not learn the negative correlation between global context features and attention distribution. We also find that the LCT block with two operators achieves the best performance, which indicates that two operators are complementary and indispensable for global context modeling.
\subsubsection{Initialization}
Table \ref{tab:init} shows the ablation results of initialization. Different from the initialization in GN, IN and LN, we find that it is more appropriate to initialize $\mathbf{w}$ and $\mathbf{b}$ to $0$ and $1$ respectively, which is consistent with the finding in SGE \cite{li2019spatial}. Initializing $\mathbf{w}$ and $\mathbf{b}$ to $0$ gets suboptimal results. As shown in Fig. \ref{fig:attention_lct_all}, we observe that most of the attention values fluctuate around $0.5$ for both SE and LCT. Hence a possible explanation is that initializing $\mathbf{w}$ to $0$ makes $\sigma(0\sim1)$ around $0.5$, which is conducive to the learning of attention distribution. When $\mathbf{w}=1$ and $\mathbf{b}=0$, LCT achieves the worst results, because the transform operator is designed to transform the context features rather than compensate for the lost of representational ability caused by normalization. 

\subsection{Object detection and segmentation on COCO}
In this section, we evaluate our block with object detection and instance segmentation tasks on the COCO-2017 dataset \cite{lin2014microsoft}. We train using 118k train images and evaluate on 5k val images. The COCO-style average precisions at different boxes and the mask IoUs are reported.
\subsubsection{Implementation details}   
All experiments are implemented with $mmdetection$ framework \cite{mmdetection}. The input images are resized such that the long edge and short edge are $1333$ and $800$ pixels respectively. We train on $4$ GPUs with $1$ images per each for $12$ epochs. All models are trained using synchronized SGD with a weight decay of $1e$-$4$ and momentum of $0.9$. According to the linear scaling rule \cite{goyal2017accurate}, the initial learning rate is set to $0.005$, which is decreased by $10$ at the $9$th and $12$th epochs. The backbones of all models are pretrained on ImageNet. We finetune all layers except for c1 and c2 with FPN \cite{lin2017feature}, detection and segmentation heads. During finetuning the BathNorm layers are frozen. Other hyper-parameters follow the default settings of the $mmdetection$ framework. The backbone is ResNet101 in all experiments.

\subsubsection{Object detection} We evaluate the LCT block on the object detection task. To this end, we insert LCT into four state-of-the-art detection frameworks, including Faster RCNN \cite{ren2015faster}, Mask RCNN \cite{he2017mask}, Cascade RCNN \cite{cai2018cascade} and Cascade Mask RCNN \cite{mmdetection}. The results on val set are given in Table \ref{tab:exp-COCO-detec}. We observe that our approach is better than SE with fewer parameters and less computations, irrespective of different detectors, which indicates that modeling global context for each channel independently is also effective on the task of object detection. In addition, compared to the baselines, LCT consistently yields $1.5\sim1.6$\% AP$_{0.5:0.95}^{bbox}$ points with neglectable extra parameters and computations, suggesting that our approach is widely applicable across various detector architectures. We also find that LCT greatly improves the detection performance of Faster RCNN, Mask RCNN and Cascade RCNN for small objects with the gain exceeding $1.9$\% AP$_{small}^{mask}$. For Cascade Mask RCNN, the detection performance of large objects is significantly boosted ($2.4\%\uparrow$ AP$_{large}^{mask}$). 
\subsubsection{Instance segmentation} 
Finally, we explore the applicability to the instance segmentation task. We select two popular frameworks, Mask RCNN and Cascade Mask RCNN. As can been seen in Table \ref{tab:exp-seg}, LCT also outperforms SE, which is consistent with the results in image classification and object detection. When adopting stronger detector Cascade Mask RCNN, the improvements achieved by LCT are still significant, suggesting that our approach is complementary to the capacity of current model. Compared to the baselines, the LCT block can boost performance by $1.1$ \% AP$_{0.5:0.95}^{mask}$ regardless of the strength of the detectors. These results suggest the generalization and effectiveness of our approach.
\section{Conclusion}
In this paper, we presented an empirical study of the relationship between global context and attention distribution of SENet. Then we considered the question of how to effectively learn the correlation between them. To this end, we introduced a simple yet effective channel attention architecture, the LCT block, to explore this question and provided experimental evidence that demonstrates the effectiveness and generalization of our approach across multiple visual tasks.  In further work, we plan to develop more efficient algorithms to exploit feature context, which may provide new insights into channel attention mechanism. 

\section{Acknowledgement}
This work is supported by the Zhejiang Provincial Natural Science Foundation (LR19F020005), National Natural Science Foundation of China (61572433, 61972347), 13-5 State S\&T Projects of China (2018ZX1030206) and thanks for a gift grant from Baidu inc.

\small
\bibliographystyle{aaai}
\bibliography{AAAI-RuanD.2195}

\end{document}